\documentclass{article}

 \usepackage[preprint]{neurips_2025}

\usepackage[english]{babel}
\usepackage[utf8]{inputenc} 
\usepackage[T1]{fontenc}    
\usepackage{hyperref}       
\usepackage{url}            
\usepackage{booktabs}       
\usepackage{amsfonts}       
\usepackage{nicefrac}       
\usepackage{microtype}      
\usepackage{xcolor}         
\usepackage{amsmath}
\usepackage{graphicx}
\usepackage{natbib}
\usepackage{amsthm}  
\usepackage{setspace} 
\usepackage{siunitx}  %
\usepackage{multirow} %
\usepackage{tabularx} 

\usepackage{makecell} %
\usepackage{algorithm}
\usepackage{algpseudocode}
\usepackage{amsmath,amssymb}
\usepackage{wrapfig}
\usepackage{float}
\usepackage{subcaption}
\usepackage{bm}

\title{MPRM: A Markov Path Rule Miner for Efficient and Interpretable Knowledge Graph Completion}
\author{%
 Mingyang Li$^{1}$, Song Wang$^{1}$, Ning Cai$^{1}$\thanks{$^{1}$School of Intelligent Engineering and Automation, Beijing University of Posts and Telecommunications, Beijing, Haidian, China. Email: \{2024010483@bupt.cn, wongsang@bupt.edu.cn, caining91@tsinghua.org.cn\}}%
}

\begin{document}
\footnotetext{This work was supported by the National Natural Science Foundation of China (Grant Nos. 62206027).}
\maketitle

\begin{abstract}
Rule mining in knowledge graphs enables interpretable link prediction. However, deep learning-based rule mining methods face significant memory and time challenges for large-scale knowledge graphs, whereas traditional approaches, limited by rigid confidence metrics, incur high computational costs despite sampling techniques. To address these challenges, we propose MPRM, a novel rule mining method that models rule-based inference as a Markov chain and uses an efficient confidence metric derived from aggregated path probabilities, significantly lowering computational demands. Experiments on multiple datasets show that MPRM efficiently mines knowledge graphs with over a million facts, sampling less than 1\% of facts on a single CPU in 22 seconds, while preserving interpretability and boosting inference accuracy by up to 11\% over baselines.
\end{abstract}

\section{Introduction}

Knowledge graphs represent real-world knowledge as triples of the form (subject, relation, object), enabling the organization of multi-relational data for applications such as search engines~\cite{SearchEngine}, recommendation systems~\cite{recommend16}, and question-answering platforms~\cite{queryanswering23,overview, KG21S}. Knowledge graph completion, also known as link prediction, is a key challenge in knowledge graph research, aiming to infer missing facts from existing ones~\cite{KNOWFORMER}. Despite recent advances, scalability and interpretability remain critical challenges in achieving effective knowledge graph completion~\cite{REVIEW23, Reasoningoverview}.

Embedding-based methods, prized for their robust predictive performance, have garnered substantial interest in recent years~\cite{RotatE, GoldE, em25}. These methods produce embeddings for entities and relations, enabling effective application in downstream tasks. While embedding-based methods excel in predictive performance, they often lack interpretability, making it challenging to provide transparent explanations required for explainable systems~\cite{Relation-Symmetrical24}. Conversely, path-based methods model reasoning as a graph search but face exponential path growth with increasing path lengths~\cite{relation10, 18variational}. To mitigate this, reinforcement learning-based methods~\cite{DEEPPATH, M-WALK, multihop, CURL} frame the search as a Markov decision process, yet struggle with sparse rewards and lengthy training times. Recent approaches, such as~\cite{RED-GNN, NBFNET, ANET}, inspired by the Bellman-Ford algorithm~\cite{bellman58}, lower complexity from exponential to polynomial through path iteration. Nevertheless, these approaches still demand significant memory. Moreover, while path-based methods provide explainable reasoning, they often fail to derive generalizable rules that yield deeper insights into the knowledge graph~\cite{Recommend}.

Unlike path-based methods, rule-based algorithms provide transparent and explicit reasoning. For example, the fact $(x, \text{father}, y)$ can be inferred from the rule $\text{mother}(x, z) \land \text{husband}(z, y) \Rightarrow \text{father}(x, y)$. This transparency is critical for reliable systems, especially in high-stakes applications. Beyond transparency, rules uncover novel patterns and knowledge within the graph~\cite{galarraga-2013-amie}. Deep learning-based rule mining methods~\cite{DRUM, NeuralLP} have been explored, but their high memory and computational demands lead to poor performance on large-scale knowledge graphs. Traditional rule mining methods often generate rules with specific constants, such as $\text{isMarriedTo}(\text{Brad Pitt}, x) \Rightarrow \text{WonPrize}(x, \text{AcademyAwards})$. Although these rules can infer facts in specific contexts, their dependence on particular entities restricts their generalizability and applicability. Additionally, these rules may lack interpretability, as they often reflect coincidental patterns rather than universal principles. Moreover, computing confidence involves frequent entity pair lookups, with each rule’s confidence calculated independently, resulting in significant computational overhead.

These limitations underscore the need for a method that combines computational efficiency with transparent, interpretable predictions. To address this gap, we introduce the Markov Path Rule Miner (MPRM), a novel approach for interpretable link prediction in knowledge graphs. Inspired by resource allocation algorithms~\cite{2007Bipartite}, MPRM reframes resource allocation as probability propagation, modeling rule-based reasoning as a Markov chain~\cite{norris1998markov}. We define confidence as the expected probability of correctly answering a query, seamlessly integrating rule mining and confidence computation with minimal overhead. Empirical evaluations demonstrate that, despite using only connected and closed Horn rules, MPRM achieves prediction performance comparable to state-of-the-art methods. By exploiting the sparsity of knowledge graphs, MPRM scales efficiently to large datasets, processing graphs with millions of facts on a standard laptop.
Our primary contributions are as follows:
\begin{itemize}
	\item We model rule-based reasoning as a Markov chain and propose a novel confidence definition based on path probability aggregation, which is computationally efficient and offers clear intuitive meaning.
	\item We present the MPRM framework, which employs a concise set of closed and connected Horn rules to deliver performance comparable to state-of-the-art methods across multiple datasets, with enhanced interpretability.
	\item For example, on the YAGO3-10 knowledge graph, with over a million facts, MPRM samples only 0.3\% of facts, achieving optimal prediction performance in 22 seconds on a single CPU. \textbf{Code is available at} \url{https://anonymous.4open.science/r/MPRM-2052/}.
\end{itemize}

\section{Preliminary}
\subsection{Horn rules and Confidence metric}
\paragraph{Problem Formulation}A knowledge graph is a set of facts $\mathcal{G} = \left\{ (s, r, o) \mid s, o \in \mathcal{E},\ r \in \mathcal{R} \right\}$, where $\mathcal{E}$ denotes the set of entities and $\mathcal{R}$ represents the set of relations. Knowledge graph completion seeks to answer queries of the form $(s,r,?)$ or $(?,r,o)$, identifying the missing entity from  $\mathcal{E}$. In this paper, we focus on queries of the form $(s,r,?)$.

\paragraph{Binary Horn Rules} A fact $(s,r,o)$ is represented as a logical atom $r(s, o)$, where $r$ is the predicate, $s$ is the subject and $o$ is the object. These atoms, true or false based on the knowledge graph, form the basis for logical rules. Formally, a \textit{binary Horn rule} is defined as:
\begin{equation}\label{eq:HornRule}
R: \quad r_0(x,z_1) \wedge  r_1(z_1,z_2) \wedge \cdots \wedge r_{T-1}(z_{T-1},y) \implies r(x,y)
\end{equation}

Equation~\ref{eq:HornRule} can be compactly expressed as $b(R) \Rightarrow h(R)$, where $h(R)$ and $b(R)$ denote the \textit{head} and \textit{body} of rule $R$. The variables $x,y,z_1,...,z_{T-1}$ can be instantiated as entities in $\mathcal{E}$, $T \in \mathbb{N}$ defining the length of rule, and each $r_m \in \mathcal{R}$ for $m=0$ to $T-1$. A rule is \textit{connected} if every atom shares at least one variable with another atom, and \textit{closed} if each variable appears in at least two atoms. We focus on mining connected and closed rules to enable interpretable knowledge graph completion.

\paragraph{Rule Confidence} In rule mining frameworks, confidence serves as a critical metric for evaluating the quality of a rule. As established in prior work~\cite{ galarraga-2013-amie,meilicke-2019-rulen}, the confidence of a rule $R$, denoted $Conf(R)$, is formally defined as:

\begin{equation}
    Conf(R) = \frac{\big|\{(x, y) \mid h(R) \land b(R) \}\big|}{\big|\{(x, y) \mid b(R)\}\big|}
    \label{eq:confidence}
\end{equation}
where $(x, y)$ represents a pair of entities. In essence, it measures the reliability of the rule by indicating how often the head holds true when the body is satisfied. However, direct confidence computation is computationally expensive due to the exponential growth of entity pairs. (i.e., the exponential increase in possible entity pairs about $|\mathcal{E}|$). While prior work~\cite{AnyBURL2019,meilicke-2019-rulen} employs sampling to estimate confidence: sacrificing accuracy for computational tractability. We fundamentally address these limitations by introducing a novel confidence metric based on Markov chain model.

\subsection{Markov-Based Confidence Measure}

Network resource allocation mechanisms~\cite{2015ModelingRP,2019RuleGuided} were originally designed for personalized recommendation systems~\cite{2007Bipartite}, our computational framework adapts this paradigm to probabilistic reasoning over knowledge graphs. Specifically, we reformulate the inference process as a Markov chain, where probability propagation replaces the concept of physical resource allocation. Specifically, we model the reasoning process using rule $R$ as a time-inhomogeneous Markov chain $(\mathcal{S},\mathcal{P})$, where:

\paragraph{State Space} $\mathcal{S} = \mathcal{E} \cup \{s_{\mathrm{abs}}\}$, where
$s_t$ is current state at time step $t$, 
$s_{\mathrm{abs}}$ is an absorbing state indicating that no further transitions to entities are possible (i.e., no reachable entities exist), and  
$s_0$ is the initial state (the query subject), representing the starting point for traversing the graph according to the rule. The chain evolves over $T$ steps, corresponding to the number of relations in the rule’s body $b(R)$.

\begin{figure}[htbp]
	\centering
	\small
	\begin{subfigure}[b]{0.40\textwidth}
		\includegraphics[width=\textwidth]{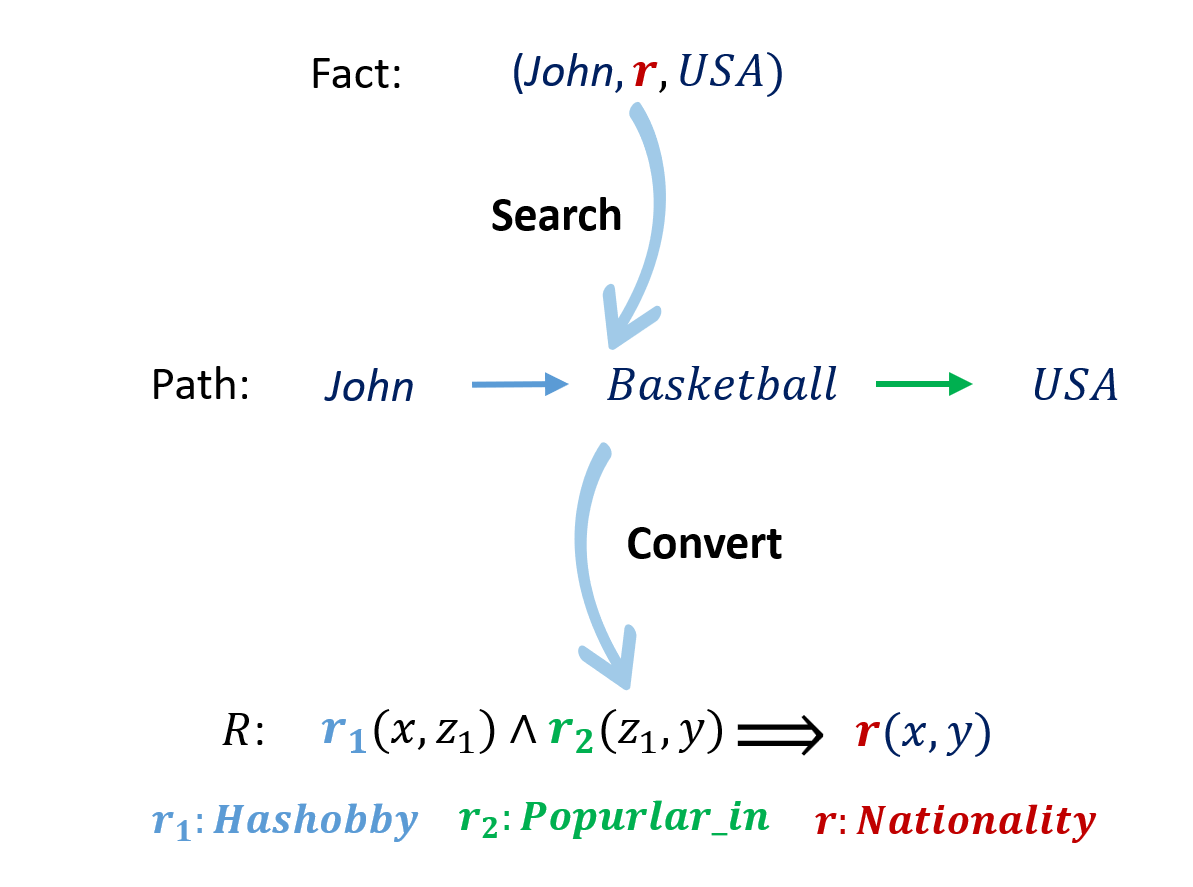}
		\caption{Rule Extraction}
		\label{fig:ruleextraction}
	\end{subfigure}
	\hfill 
	\begin{subfigure}[b]{0.40\textwidth}
		\includegraphics[width=\textwidth]{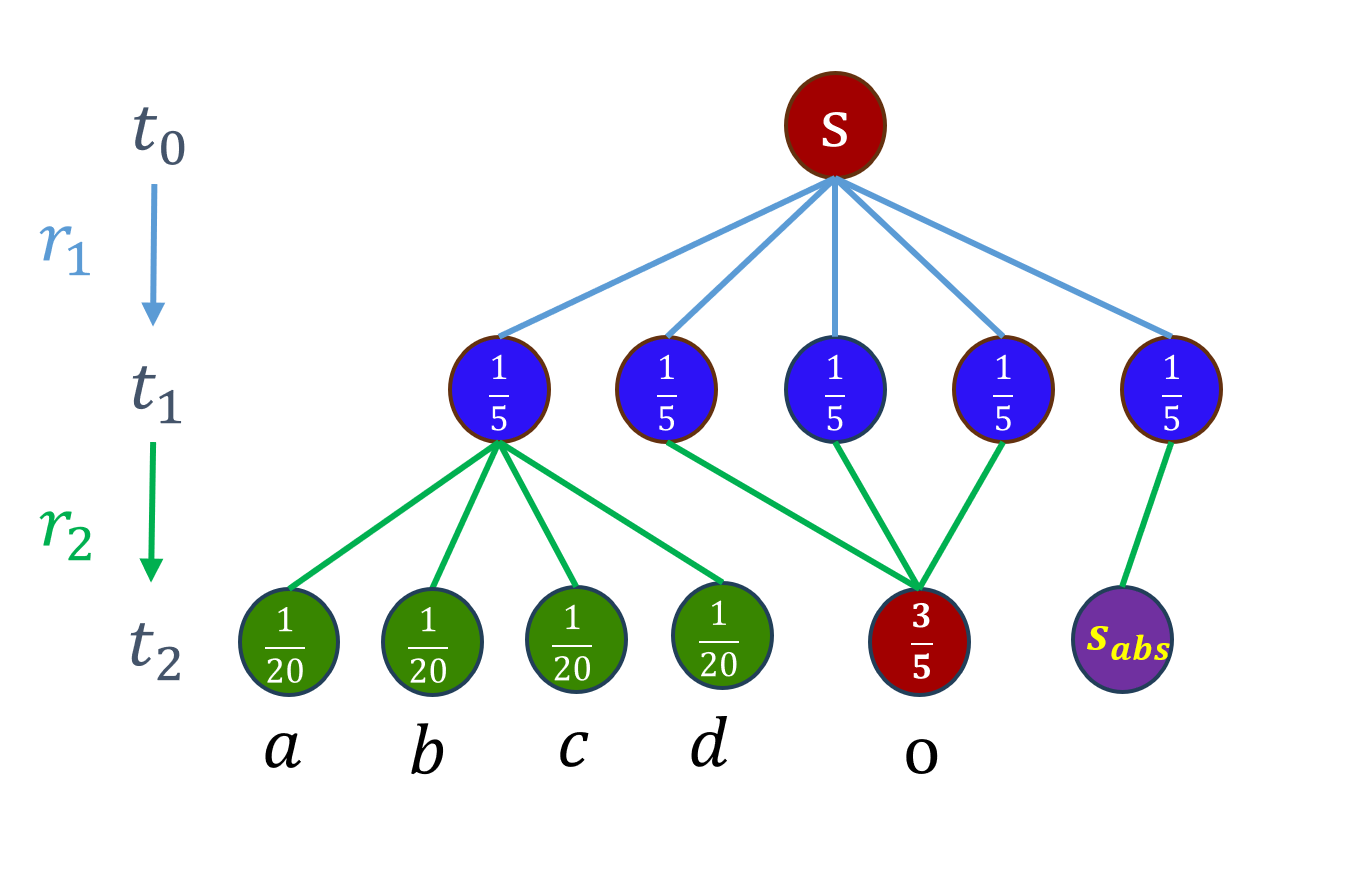}
		\caption{Rule-based Reasoning}
		\label{fig:ruleapplication}
	\end{subfigure}
	\caption{Rule extraction and reasoning framework. The extracted rule $r_1(x, z_1) $ $\land r_2(z_1, y)$  $\implies r(x, y)$ resolves the query $(s,r,?)$, where $Q(s,r)=\{o\}$. The numerical values annotated on each node represent the probability of reaching the corresponding node at time step $t$. }
	\label{fig:total}
\end{figure}

\paragraph{Transition Probabilities} For each step $t$ from 0 to $T-1$, the transition probabilities are defined based on the fixed sequence of relations $(r_0,r_1,...,r_{T-1})$ in $b(R)$. Specifically, the probability of transitioning from state $s_t$ to $s_{t+1}$ is:
\begin{equation}
\mathcal{P}(s_{t+1} \mid s_t, r_t) = 
\begin{cases}
\frac{1}{|Q(s_t, r_t)|} & \text{if } s_{t} \neq s_{\mathrm{abs}}\land s_{t+1} \in Q(s_t, r_t) \\
1 & \text{if } s_{t} \neq s_{\mathrm{abs}}\land  Q(s_t, r_t)= \emptyset\land s_{t+1}=s_{\mathrm{abs}}  \\
1 & \text{if } s_{t} = s_{\mathrm{abs}} \land s_{t+1} = s_{\mathrm{abs}}\\
0 & \text{otherwise}
\end{cases}
\end{equation}
where $Q(s_t, r_t) = \{ o \mid (s_t, r_t, o) \in \mathcal{G} \}$ denotes the set of reachable entities from $s_t$ via the relation $r_t$ at step $t$, and  $|Q(s_t, r_t)|$ is the number of such entities. This model simulates the process of traversing the knowledge graph following the predetermined sequence of relations in the rule’s body. At each step $t$,  the chain moves to a new entity based on $r_t$, with equal probability among possible next entities  (indicating no prior knowledge about entity preferences), or transitions to $s_{\mathrm{abs}}$ if no entities are reachable. Once in $s_{\mathrm{abs}}$, the chain remains there, capturing cases where the reasoning cannot be completed. After $T$ steps, the final state $s_T$ represents a potential answer to the query, or $s_{\mathrm{abs}}$ if the reasoning fails. Figure~\ref{fig:ruleapplication} is an instance of this process.
\subsection{Rule Confidence via Path Probability Aggregation}
 Building on the Markov chain model, we define rule confidence as the expected probability of correctly answering queries through rule-guided inference. Under the Markov property, the path probability for a transition sequence
$\xi: s_0 \xrightarrow{r_0} \cdots \xrightarrow{r_{T-1}} s_T$ is:
\begin{equation}\label{eq:PathP}
P(\xi \mid R) = \prod_{t=0}^{T-1} \mathcal{P}(s_{t+1} \mid s_t, r_t)
\end{equation}
Let $ \Xi_{s\rightarrow o}^{R}$ denote  the  set  of  all  simple  paths  from  $s$  to  $o$  under  rule  $R$.  The  probability  of  reaching  $o$  from  $s$  is  defined  as:
\begin{equation}
P(s_T = o\mid s_0=s,R) = \sum_{\xi \in \Xi_{s\rightarrow o}^{R}} P(\xi \mid R)
\label{eq:total_prob}
\end{equation}
As queries may have multiple correct answers, rules that reliably infer more answers should have stronger confidence. To capture this, we introduce the Path Reachability Mass (PRM) as the summation of probabilities for reaching all valid answers. Formally, for a rule 
$R$ and query subject $s$, the PRM is given by:
\begin{equation}
\mathit{PRM}(R, s) = \sum_{o \in Q(s,r)} P(s_T = o \mid s_0 = s, R)
\label{eq:prm}
\end{equation}
Building on this measure, the ultimate confidence of rule $R$ is subsequently defined through expectation over the queries with relation $r$:
\begin{equation}
\label{eq:conf2}
 \mathit{PConf}(R) = \mathbb{E}_{(s,r,?)\sim \mathcal{G}_{r}} PRM(R,s)
\end{equation}
 where $\mathcal{G}_{r}$ is the set of facts with relation $r$ in $\mathcal{G}$. This confidence metric quantifies  the expected probability that rule $R$ correctly answers the queries. The definition of $Conf$ can result in confidence of 1 even when only a single entity pair satisfies both the head and body of a rule. In contrast, $PConf$ addresses this issue by calculating confidence based on the entire set of facts to the predicate of the rule's head, ensuring that high-confidence rules are applicable to most queries.

\paragraph{Mechanism Analysis } The traditional confidence measure (Equation~\ref{eq:confidence}) quantifies the proportion of entity pairs \( (s, o) \) where the rule’s body and head both hold. However, this metric ignores the varying probabilities of reaching different entities, treating all valid pairs equally. Figure~\ref{fig:ruleapplication} shows that the probability of reaching entity $y$ far exceeds that of other nodes, yet the traditional metric (Equation~\ref{eq:confidence}) treats all pairs equally, ignoring these differences. Moreover, $PConf$ prioritizes rules with N-to-1 or 1-to-1 relations in their body, which yield higher path probabilities and thus increase rule confidence. These relations also reduce the search space, enabling more deterministic transitions and fewer paths. These dual advantages—\textbf{sensitivity to path probabilities} and \textbf{specificity from constrained relations}—drive $PConf$’s superior performance over traditional confidence measures.

\section{Markov Path-based Rule Miner}
This section introduces the proposed rule mining algorithm  MPRM. Notably, the confidence calculation presented in Equation~\ref{eq:conf2} offers a simpler approach. This simplicity stems from the fact that the path probability is computed as the product of $\frac{1}{|Q(s,r)|}$, where $|Q(s,r)|$ can be efficiently retrieved through a lookup table with a time complexity of $\mathcal{O}(1)$.
\subsection{Rule Extraction}
\label{sec:framework}
Given a knowledge graph $\mathcal{G}$ augmented with inverse facts (for each fact $(s,r,o)$, add an inverse fact $(o,r^{-1},s)$ ),  MPRM operates through three steps:

 \textbf{Step 1: Single-hop Rule Extraction} 
For each fact $(s, r, o) \in \mathcal{G}$, we identify all relations $r_j \neq r$ such that $(s, r_j, o) \in \mathcal{G}$ and generate length-1 rules $R_{sg}: r_j(x,y) \Rightarrow r(x,y)$, capturing  implications where $r_j$ between entities implies $r$. For those length-1 rules the PRM is computed efficiently using the following formula:
\begin{equation}
    PRM(R_{sg},s) = \frac{|Q(s, r) \cap Q(s, r_j)|}{|Q(s, r_j)|}
    \label{eq:single_hop}
\end{equation}

\textbf{Step 2: Multi-hop Rule Extraction}
For each fact $(s, r,o)\in \mathcal{G}$, we execute Bidirectional Breadth-First Search (Bi-BFS) to discover all simple paths $\{\Xi_{s\rightarrow o'}\}_{o' \in Q(s,r)}$ with path lengths $l$ (number of edges) satisfying $2 \leq l \leq L$, where $L$ is the maximum rule length.
Each discovered path $\xi: s \xrightarrow{r_0} s_1 \xrightarrow{r_1} \cdots \xrightarrow{r_{l-1}} s_{l}$ is mapped into a length-$l$ Horn rule as shown in Figure~\ref{fig:ruleextraction}. For each fact $(s, r,o)$ the PRM is computed based on rule instantiation as follows: if the rule $R$  is successfully instantiated during path exploration (i.e., if the path matches the rule's body), PRM is computed via Equation~\ref{eq:prm}; otherwise, PRM equals zero.

\textbf{Step 3: Confidence normalization} The confidence metric, as defined in Equation~\ref{eq:conf2}, is computed by aggregating all $PRM$ values associated with a given rule $R$. This aggregated value is then normalized by the number of facts in the knowledge graph $\mathcal{G}$ with relation $h(R)$.

\subsection{Reasoning}
We describe how mined rules are applied to knowledge graph completion for queries of the form $(s, r_i,?)$. We use rules from the set $R_i=\{R|h(R)=r_i(x,y)\}$, however, applying all such rules is computationally prohibitive. We select the top $K$ rules from $R_i$, ranked by their confidence, denoted $TopK(R_i)$. For each candidate entity $y$, we compute a score by aggregating contributions from rules in $TopK(R_i)$, where each rule $R$'s contribution is 
$P(s_T=y|s_0=s,R) \cdot PConf(R)$. Two methods are available for performing this aggregation:
\begin{equation}
s_{sum}(y)=\sum_{R \in TopK(R_i)}  [P(s_T=y|s_0=s,R) \cdot PConf(R)]
\label{eq:predict_sum}
\end{equation}
\begin{equation}
s_{max}(y)=\max_{R \in TopK(R_i)}[P(s_T=y|s_0=s,R)\cdot PConf(R)]
\label{eq:predict_max}
\end{equation}
Since the accuracy of both methods is similar, we only report the results of $s_{sum}$ for simplicity. 

\subsection{Time Complexity Analysis}

\begin{minipage}{0.37\textwidth}
	The time complexity of MPRM stems from its multi-hop rule extraction, as detailed in Algorithm~\ref{alg:my_algorithm}.  Bi-BFS maintains forward and backward queues. The forward queue starts with the source node $s$, while the backward queue contains all possible answers in \( Q(s, r) \). Bi-BFS efficiently identifies all simple paths and aggregates their path probabilities to compute rule confidence. The theoretical complexity of MPRM is bounded by $\mathcal{O}(\alpha |\mathcal{R}||Q| d^{\lceil L/2\rceil})$, where $|Q|$ is the average number of answers (typically small in real-world knowledge graphs), $d$ is the average degree, $\alpha$ is a parameter controlling the maximum number of triples per relation and $L$ is the maximum rule length. Experimental results demonstrate that setting $\alpha$ to 100 achieves performance comparable to full graph traversal 
\end{minipage}
\hfill
\begin{minipage}{0.6\textwidth}
	\begin{algorithm}[H]
		\caption{Multi-hop Rule Extraction}\label{alg:my_algorithm}
		\textbf{Input:} 
		Knowledge Graph: $\mathcal{G}$, Maximum Rule Length: $L$, $\alpha$\\
		\textbf{Output:} $Rules$  and $PConf$ \\
		\textbf{Initialize:}  $Rules  \gets \emptyset, PConf  \gets \{\} $
		\begin{spacing}{1.1}
			\begin{algorithmic}[1]
				\State Convert $\mathcal{G}$ to undirected graph $G$ 
				\State $\mathcal{G'}\gets$ $Sample(\mathcal{G},\alpha)$
				\For {$(s,r,o) \in \mathcal{G'}$}
				\State $\Xi_{s\rightarrow Q(s,r)} \gets \text{Bi-BFS}(G,s,\{Q(s,r)\}, L)$
				\For {$\xi \in \Xi_{s\rightarrow Q(s,r)}$}
				\State $Rule\gets $Convert $\xi$ to $Rule$
				\State $P(\xi|Rule)\gets$Compute Path prob.by Eq~\ref{eq:PathP}
				\State $PConf[Rule] \gets PConf[Rule]+P(\xi|Rule)$
				\State $Rules \gets \cup Rule$
				\EndFor
				\EndFor
				\State $PConf \gets Normalize(PConf)$ 
				\State \Return $Rules, PConf$ 
			\end{algorithmic}
		\end{spacing}
	\end{algorithm}
	\vspace{1em} 
\end{minipage}
(as shown in Section~\ref{sec:ablation}).

For large-scale, real-world knowledge graphs like Wikidata or YAGO, which contain over $10^8$ entities but exhibit sparse connectivity $(d<30)$, this approach proves highly efficient. For instance, in 6-hop rule mining (where $L=6$ and $\lceil L/2\rceil=3$), the number of operations per fact is approximately $|Q|\cdot d^3\approx 10^5$, three orders of magnitude below the entity count $|\mathcal{E}|$. Our method remains scalable across any knowledge graph, provided the number of relation types $|\mathcal{R}|$ stays bounded.

\section{Experiments}
\subsection{Experiment Setup and Main Result}
\label{sec:Experiment details}
\paragraph{Datasets \& Evaluation}
We evaluate MPRM on four standard datasets: FB15K-237~\cite{FB15K-237}, WN18RR~\cite{ConvE}, NELL-995~\cite{DEEPPATH}, and YAGO3-10~\cite{YAGO3-10}. Dataset statistics are shown in Table~\ref{tab:datasetdetail}. We use original splits from the dataset papers. We use filtered metrics~\cite{Survey} Mean Reciprocal Rank (MRR), Hit@1, and Hit@10, with higher values indicating better performance.

\paragraph{Baselines}
We compare MPRM against several approaches: rule mining methods (AMIE~\cite{galarraga-2013-amie}, AnyBURL~\cite{AnyBURL2019}, RuleN~\cite{meilicke-2019-rulen}, NeuralLP~\cite{NeuralLP}, DRUM~\cite{DRUM}),  reinforcement learning-based and path-based methods (MINERVA~\cite{MINERVA}, CURL~\cite{CURL}, A*NET~\cite{ANET}), and embedding-based methods (TransE~\cite{TransE}, RotatE~\cite{RotatE}, HousE~\cite{house}).

\begin{table}[t]
	\centering
	\caption{Dataset statistics for link prediction}
	\begin{tabular}{lrrrr}
		\toprule
		\textbf{Dataset} & {$|\mathcal{R}|$} & $|\mathcal{E}|$ & $|\mathcal{G}|$ & Test \\
		\midrule
		\textbf{FB15K-237} & 237 & 14,541 & 272,115 & 20,466 \\
		\textbf{WN18RR} & 11 & 40,559 & 86,835 & 2,924 \\
		\textbf{YAGO3-10} & 37 & 123,182 & 1,079,040 & 5,000 \\
		\textbf{NELL-995} & 200 &74,432 & 149,678 & 2,818 \\
		\bottomrule
		\label{tab:datasetdetail}
	\end{tabular}
\end{table}

\begin{table}[ht]
	\centering
	\caption{Link prediction performance on four standard datasets. The best results are \textbf{boldfaced} and the second-best results are \underline{underlined}. A*NET is the state-of-the-art method for the FB15K-237 and WN18RR datasets among path-based methods. }
	\footnotesize  
	\begin{tabularx}{\textwidth}{c *{12}{c}}
		\toprule
		\multicolumn{1}{c}{Method} & \multicolumn{3}{c}{\textbf{FB15K-237}} & \multicolumn{3}{c}{\textbf{WN18RR}} & \multicolumn{3}{c}{\textbf{YAGO3-10}} & \multicolumn{3}{c}{\textbf{NELL-995}} \\
		\cmidrule(lr){2-4} \cmidrule(lr){5-7} \cmidrule(lr){8-10} \cmidrule(lr){11-13}
		& & \multicolumn{2}{c}{Hit} &  & \multicolumn{2}{c}{Hit} &  & \multicolumn{2}{c}{Hit} &  & \multicolumn{2}{c}{Hit} \\
		\cmidrule(lr){3-4} \cmidrule(lr){6-7} \cmidrule(lr){9-10} \cmidrule(lr){12-13}
		& MRR & @1 & @10 & MRR & @1 & @10 & MRR & @1 & @10 & MRR & @1 & @10 \\
		\midrule
		\mbox{AMIE } &.201  &- &.362  &.356   &-  &.357 &-   &- &- &-   &- &-\\
		\mbox{RuleN } & - & .182 & .420 & - & .427 & .536 &- &- &- &-   &-  &-\\
		\mbox{AnyBURL} & .310 & .233 & .486 & .480 &.446  & .555  &.540 &.477 &.673 &-   &-  &-\\
		NeuralLP  &.252 &.189  & .375    &.435   & .371    &.566    &-   &-  &-   &-   &-  &-      \\
		\mbox{DRUM }  &.343 & .255  & .516   & .486  & .425 & .586    &.531 &.453 &.676  &.532   &.460  &.662\\
		\midrule
		\mbox{TransE } & {.330} & {.232} & .526 &.222 &.014 & .528 &.510 &.413 &.681  & .507  &.424  &.648\\
		\mbox{RotatE }&.337 &.241  &.533   &.477 &.428 &.571 &.495 &.402 &.670 & .619  &.542  &  .752        \\
		\mbox{HousE } & \underline{.361} & \underline{.266} & .551 & .511 & .465 & .602 &\underline{.571} &\underline{491} &\underline{.714} &-   &-  &-\\
		\midrule
		MINERVA &.293 &.217  &.456    & .448  & .413 & .513    &-    &-    &-  &{.725}   & \underline{.663}  &{.831}\\
		\mbox{CURL }   &306 &.224   &.470  &.460   &.429  &.523  &-  &-  &- &\textbf{.738}  &\textbf{.667} &\underline{.843}      \\
		\mbox{A*NET } &\textbf{.505} & \textbf{.410}  & \textbf{.687}   & \textbf{.557}  & \textbf{.504} & \textbf{.666}    &556 &.470 &.707  &.521   &.447  &.631\\
		\midrule
		MPRM   &.351 & .234 & \underline{.556} & \underline{.537} & \underline{.478} & \underline{.632}  &\textbf{.635} &\textbf{.549} &\textbf{.778} &\textbf{.738}   &{.660}  &\textbf{.861} \\
		\bottomrule
	\end{tabularx}
	\label{tab:method_comparison}
\end{table}
\paragraph{ Experimental Details }

Knowledge graphs typically exhibit a long-tail distribution in node degrees, with a small fraction of nodes being densely connected. Naively expanding these nodes is computationally costly. To address this, we introduce a hyperparameter 
\(\beta\) to limit the number of nodes expanded per node in Bi-BFS. This is similar to a dropout mechanism in reinforcement learning algorithms~\cite{MINERVA,multihop}, randomly selecting up to \(\beta\) child nodes from a node’s neighbors. As query answers also follow a long-tail distribution, we limit the search to paths yielding up to 5 results, covering 90\% of queries. The training process thus uses three hyperparameters: $L$, $\alpha$ and $\beta$. We set $L=6$ for WN18RR and $L=3$ for other datasets. We use $\beta=100$ for all datasets except FB15K-237, where $\beta=200$ . For prediction, we use the top 300 rules ranked by confidence for each query. Experiments were conducted in Python on an Intel i9-14900H CPU, without GPUs. 

\begin{figure}[htbp]
	\centering
	\begin{subfigure}[b]{0.45\textwidth}
		\includegraphics[width=\textwidth]{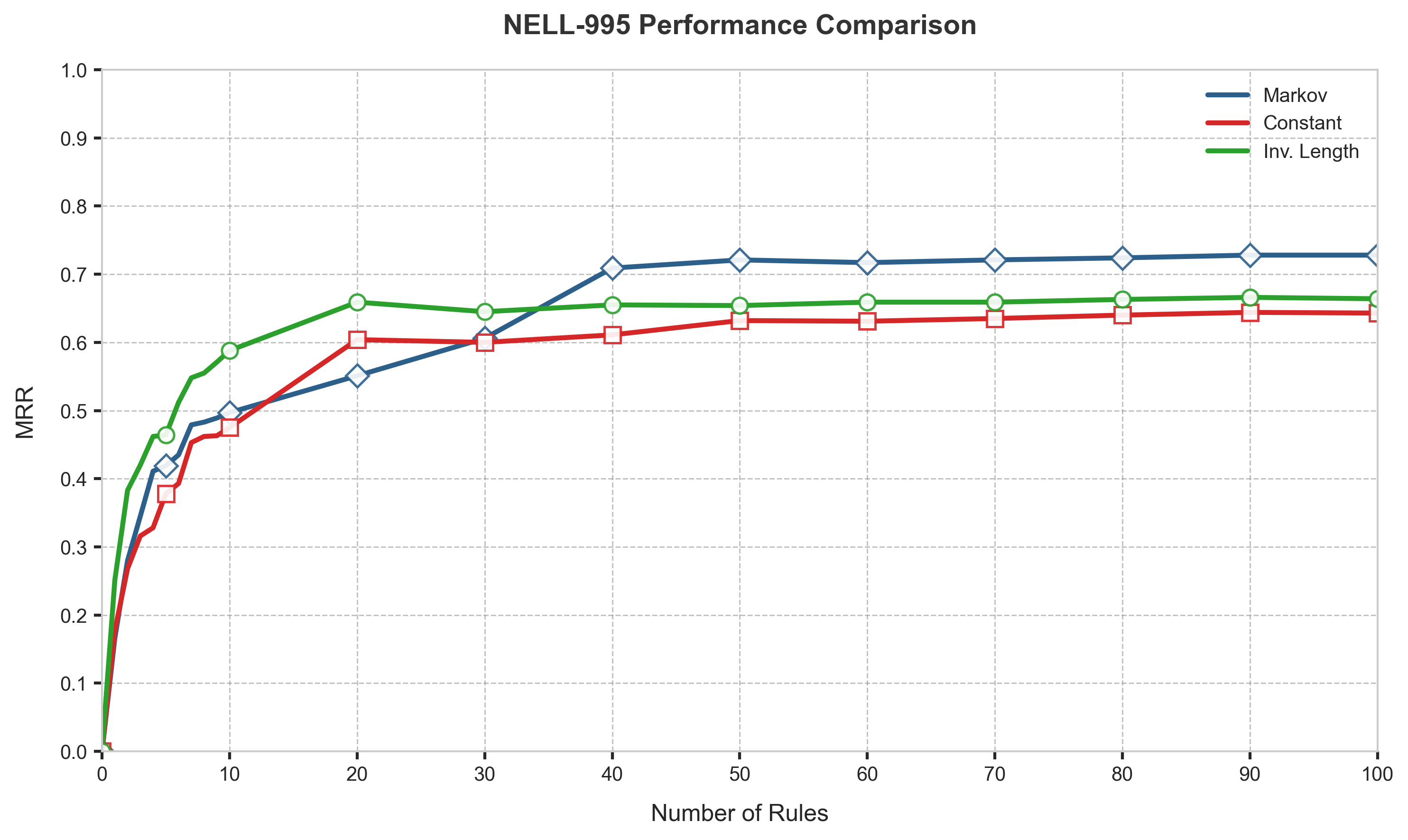}
		\caption{NELL-995}
		\label{fig:sub1}
	\end{subfigure}
	\hfill 
	\begin{subfigure}[b]{0.45\textwidth}
		\includegraphics[width=\textwidth]{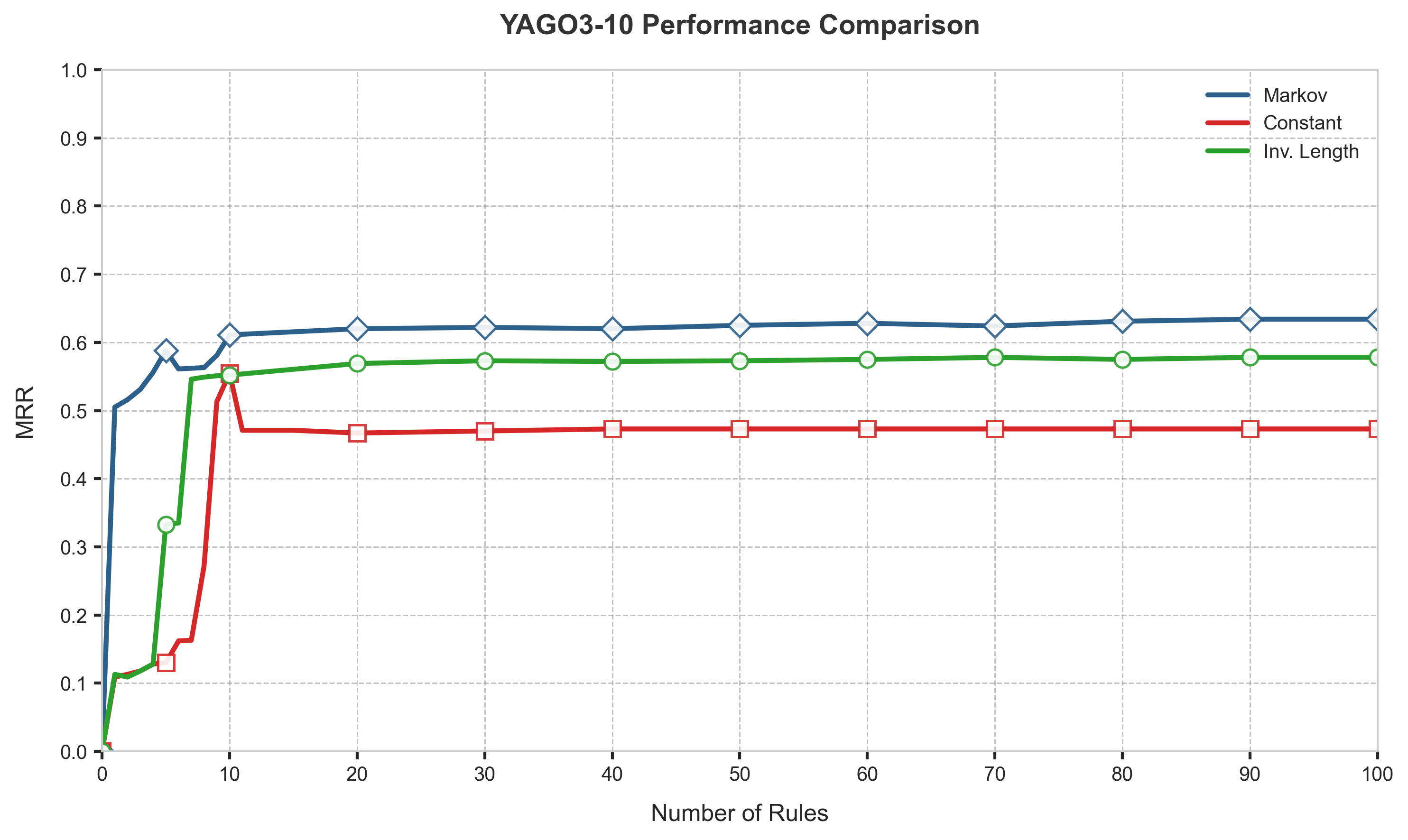}
		\caption{YAGO3-10}
		\label{fig:sub2}
	\end{subfigure}
	\caption{Ablation study on path probability.}
	\label{fig:Ablation}
\end{figure}

Results are shown in Table~\ref{tab:method_comparison}. Compared to other rule-based methods, MPRM achieves higher performance with greater computational efficiency, outperforming existing approaches. Importantly, rule-based methods like MPRM ensure full interpretability, with transparent answer derivations. Table~\ref{tab:time_comparison} reports the rule mining time for optimal performance. On YAGO3-10, MPRM achieved an 11.2\% MRR improvement over the second best baselines in only 22 seconds on a single CPU.

\begin{table}[htbp]
\centering
\caption{Running time of MPRM on four datasets. }
\label{tab:time_comparison}
\begin{tabular}{lcccc}
\toprule
\textbf{Method} & \textbf{FB15K-237} & \textbf{NELL-995} & \textbf{YAGO3-10}  & \textbf{WN18RR} \\
\midrule
MPRM     & {581s}      & {26s}     & {22s}   & {9s}  \\
\bottomrule
\end{tabular}
\vspace{0.5em}
\small
\end{table}

\subsection{Ablation studies}
\label{sec:ablation}
\paragraph{Path probability}To evaluate the Markov-based path probability measure, we compare two alternatives: (1) \textit{Length-based}, where path probability is the reciprocal of path length, and (2) \textit{Constant}, where path probability is fixed at 1, relying only on rule frequency. We conducted experiments on YAGO3-10 and NELL-995, with results shown in Figure~\ref{fig:Ablation}. The Markov-based measure significantly improves rule quality. On YAGO3-10, the Markov-based approach achieves an MRR of 0.5 with a single rule, compared to approximately 0.11 for the Length-based and Constant alternatives.

\begin{wrapfigure}{r}{0.45\textwidth}
       \centering
      \includegraphics[width=0.40\textwidth]{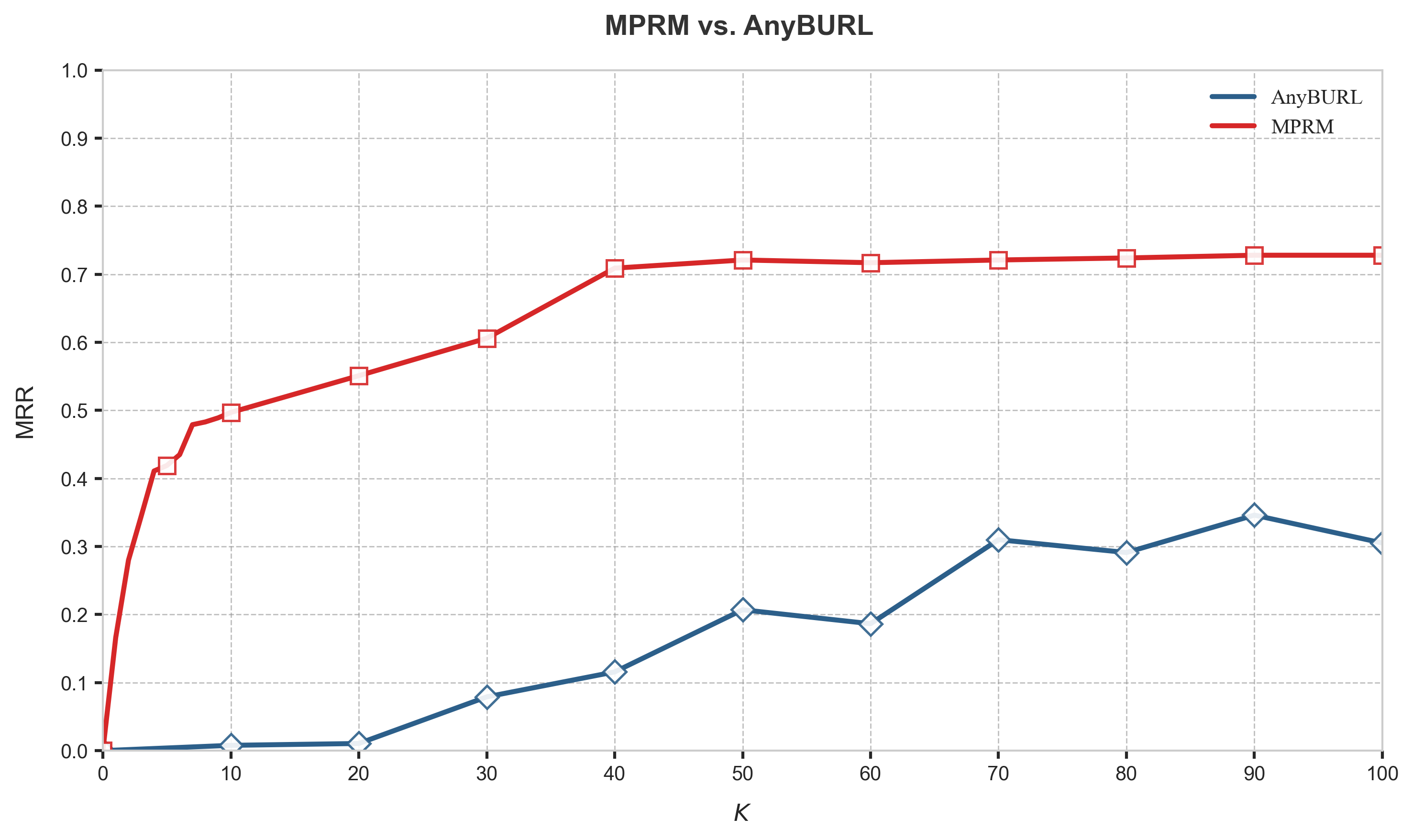}
      \caption{AnyBURL runs for 100 seconds and MPRM for 26 seconds, after which each selects the $K$ rules with the highest confidence for link prediction.}
      \label{fig:Comparison}
\end{wrapfigure}
 \paragraph{Rule Quality}
 We compare MPRM with the state-of-the-art rule mining method AnyBURL. With the same number of confidence-ranked rules, MPRM significantly outperforms AnyBURL. As shown in Figure~\ref{fig:Comparison}, MPRM achieves an MRR of 0.6 on NELL-995 using only 30 rules per query, compared to AnyBURL's MRR of 0.1. AnyBURL primarily learns rules incorporating constants, which often lack interpretability due to their tendency to overfit specific facts in the knowledge graph, rather than capturing generalizable patterns. For instance, the rule $ \text{actedIn}(x, z_1) \land \text{actedIn}(z_2, z_1) \land \text{isMarriedTo}(\text{Sandra\_Bullock}, z_2)$ $\implies$ $\text{hasGender}(x, \text{female})$ extracted by AnyBURL achieves a confidence of 0.833. However, its reliance on the specific entity "Sandra\_Bullock"  undermines its interpretability, as it fails to provide a generalizable basis for gender prediction. In contrast, MPRM learns only rules without constants, which are simpler and more interpretable. Rules shown in Table~\ref{tab:rule:visualization} align with human intuition, e.g., children and parents often win the same prize, or graduation and workplace locations coincide.

\begin{table}[h]
	\centering
	\caption{Rules learned by MPRM on the YAGO3-10 dataset.}
	\label{tab:rule:visualization}
	\scriptsize
	\begin{tabular}{p{0.05\textwidth} p{0.45\textwidth} p{0.40\textwidth}}
		\toprule
		\textbf{Head} & \textbf{WonPrize(x,y)} & \textbf{WorksAt(x,y)} \\
		\midrule
		\multirow{3}{*}{\textbf{Body}} & \( \text{WonPrize}(x,z_1)\land \text{WonPrize}(z_2,z_1)\land \text{WonPrize}(z_2,y) \) & \( \text{GraduatedFrom}(x,y) \) \\
		& \( \text{HasChild}(z_1,x) \land \text{WonPrize}(z_1,y) \) & \( \text{HasAcademicAdvisor}(z_1,x) \land \text{GraduatedFrom}(z_1,y) \) \\
		& \( \text{Influences}(x,z_1) \land \text{WonPrize}(z_1,y) \) & \( \text{HasAcademicAdvisor}(x,z_1) \land \text{WorksAt}(z_1,y) \) \\
		\bottomrule
	\end{tabular}
\end{table}

\begin{figure}[htbp]
\centering
    \includegraphics[width=0.45\textwidth]{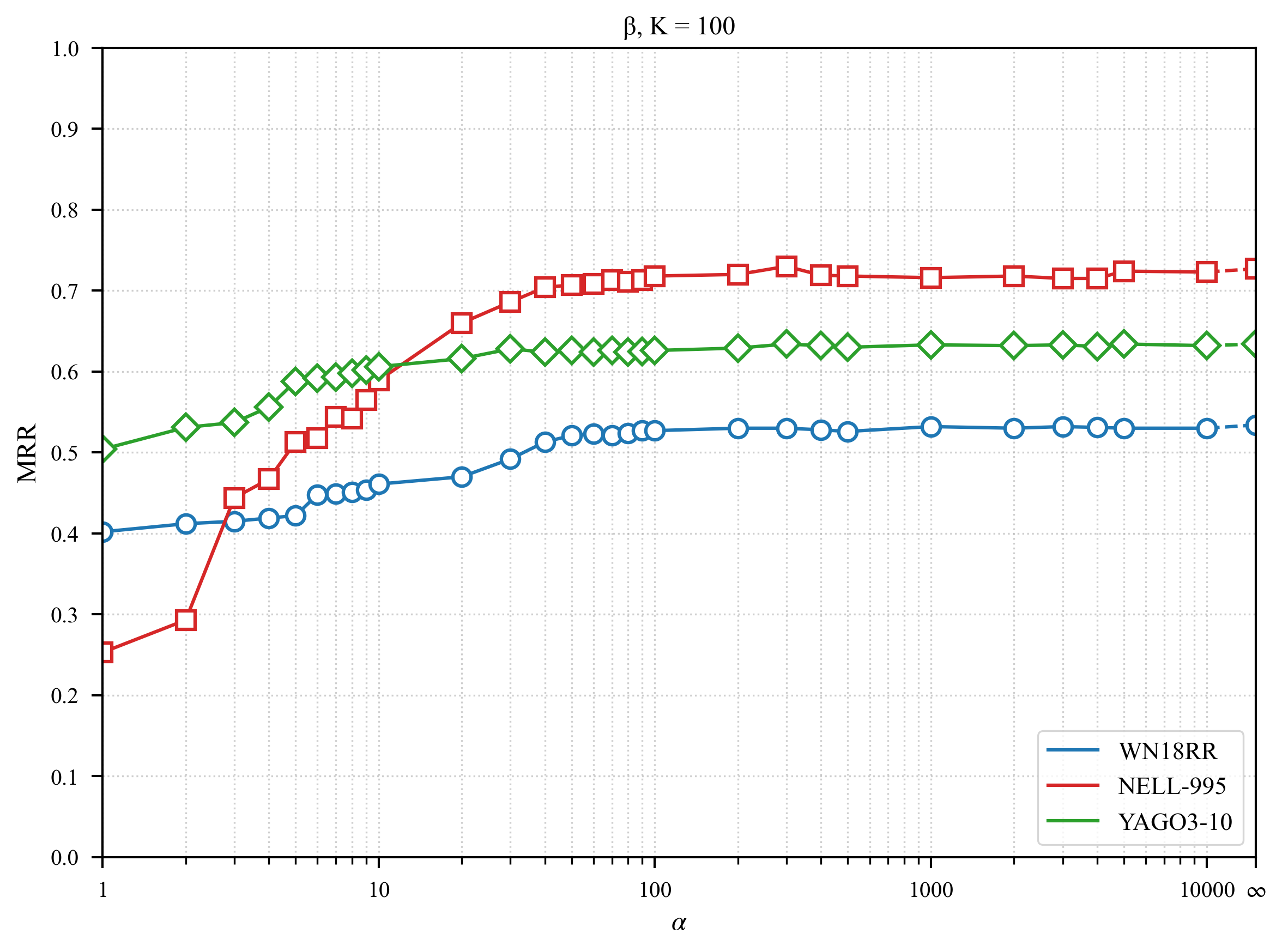}
    \caption{As $\alpha$ varies from 1 to infinity with other hyperparameter held constant. The result on three datasets show that each relation requires only a small number of facts to achieve the accuracy of learning all facts, highlighting the efficiency of MPRM.}
    \label{fig:multi_curve_mrr}
\end{figure}
\paragraph{Efficiency}
In terms of memory usage, MPRM stores only the adjacency list and mined rules. Although the number of paths grows exponentially with rule length, MPRM does not store paths; instead, it converts each path found via Bi-BFS into a rule and computes its probability immediately. In contrast, deep learning methods like DRUM~\cite{DRUM} and NeuralLP~\cite{NeuralLP} face memory overflow when processing large knowledge graphs like YAGO3-10. Moreover, MPRM learns semantic relationships between relations from a small fraction of facts. By tuning the hyperparameter \(\alpha\), which controls the number of facts sampled per relation (with \(\alpha \to \infty\) using the entire training set), MPRM achieves performance comparable to the full dateset using only 100 facts per relation—13\% of facts in NELL-995, 1.2\% in WN18RR, and 0.3\% in YAGO3-10, as shown in Figure~\ref{fig:multi_curve_mrr}.


\section{Related Works}
\paragraph{Embedding-based Methods}

Recent advancements in embedding methods, including TransE~\cite{TransE} Dist-Mult~\cite{Distmult}, ComplEx~\cite{ComplEx}, HousE~\cite{house}, and GoldE~\cite{GoldE}, enhance expressiveness by refining the loss function or embedding space. Work in~\cite{2015ModelingRP} integrates resource allocation algorithms with embedding representations, using relational path energy to improve TransE’s energy function. Additionally, work in~\cite{2019RuleGuided} uses rule confidence to inform path-based predictions. Both approaches seek to enhance embedding models but do not apply resource allocation to rule mining in knowledge graphs.

\paragraph{Rule Mining Methods}
Horn rule mining has been a focus since Inductive Logic Programming (ILP)~\cite{BLP,IEP,LIR}. However, these methods rely on negative examples, which are inherently scarce in knowledge graphs. Inspired by association rule mining~\cite{AM}, AMIE targets rule mining in Resource Description Framework (RDF) knowledge graphs. Its iterative rule generation process, however, results in incomplete rules. AMIE+~\cite{AMIE+} improves efficiency by refining metrics for rule extension and pruning, yet it does not exploit closed paths in knowledge graphs. Subsequently, RuleN~\cite{RuleE} and AnyBURL~\cite{AnyBURL2019} generate closed rules based on closed paths. These algorithms, however, depend on traditional confidence definitions, which introduce computational challenges. Moreover, they extract both closed Horn rules and rules with constants, producing a large rule set scaling with the number of entities. In contrast, DRUM~\cite{DRUM} and NeuralLP~\cite{NeuralLP} use deep learning to mine closed rules, but their high computational complexity, due to computing adjacency matrix products, yields performance inferior to traditional rule mining approaches.
\paragraph{Confidence Measures}
Due to the open-world assumption, facts absent from the graph are not necessarily false, rendering confidence calculations imprecise. Efforts to optimize confidence measures, as in~\cite{Conf18,AMIE+}, are limited to association rule frameworks, relying on entity pair proportions while ignoring probability distributions. In contrast, work in~\cite{ModifedConf} uses logistic regression to refine rule confidence, significantly improving predictive performance over traditional measures. This advancement highlights the clear need for novel confidence definitions.
\section{Discussion and Conclusion}
\label{sec:discuss}
\paragraph{Limitations} 
Hyperparameter selection is empirical rather than theoretical, and optimally adapting hyperparameters across diverse datasets remains an open challenge. Additionally, to balance efficiency and interpretability, we focus exclusively on mining closed rules. However, knowledge graphs include other rule types (e.g., open rules), and this focus limits MPRM’s reasoning capabilities. Thus, extending MPRM to include diverse rule types will be a key focus of future research.
\paragraph{Social Impacts}
MPRM’s efficiency allows its deployment on resource-constrained devices, such as smartphones, for applications like recommendation systems. However, processing sensitive user data on these devices raises privacy concerns due to their limited security capabilities. Future work will explore differential privacy techniques to address these risks.

\paragraph{Conclusion}
We propose MPRM, an efficient rule mining framework for knowledge graphs that models rule-based reasoning as a Markov chain and introduces a novel confidence measure. By leveraging only closed connected rules, MPRM achieves superior performance in knowledge graph completion, rapidly processing large-scale graphs and extracting highly interpretable rules. Experiments on multiple datasets demonstrate that MPRM outperforms most baselines in interpretability, reasoning accuracy, and efficiency, revealing the untapped potential of rule-based methods.

\bibliographystyle{plain}
\bibliography{refs}  

\newpage
\section*{NeurIPS Paper Checklist}

\begin{enumerate}

\item {\bf Claims}
    \item[] Question: Do the main claims made in the abstract and introduction accurately reflect the paper's contributions and scope?
    \item[] Answer: \answerYes{} 
    \item[] Justification: The contributions and scope mentioned in the abstract and introduction are consistent with the content of the paper. The paper indeed proposes a new rule mining method called MPRM and verifies its efficiency and interpretability through experiments.

\item {\bf Limitations}
    \item[] Question: Does the paper discuss the limitations of the work performed by the authors?
    \item[] Answer: \answerYes{} 
    \item[] Justification: We discuss the limitations of MPRM in Section~\ref{sec:discuss}.

\item {\bf Theory assumptions and proofs}
    \item[] Question: For each theoretical result, does the paper provide the full set of assumptions and a complete (and correct) proof?
    \item[] Answer: \answerNA{} 
    \item[] Justification: This article contains no theoretical proofs.

    \item {\bf Experimental result reproducibility}
    \item[] Question: Does the paper fully disclose all the information needed to reproduce the main experimental results of the paper to the extent that it affects the main claims and/or conclusions of the paper (regardless of whether the code and data are provided or not)?
    \item[] Answer:  \answerYes{} 
    \item[] Justification: All experimental details is reported in Section~\ref{sec:Experiment details}. 

\item {\bf Open access to data and code}
    \item[] Question: Does the paper provide open access to the data and code, with sufficient instructions to faithfully reproduce the main experimental results, as described in supplemental material?
    \item[] Answer: \answerYes{} 
    \item[] Justification: Data and code are available at \url{https://anonymous.4open.science/r/MPRM-2052/}.

\item {\bf Experimental setting/details}
    \item[] Question: Does the paper specify all the training and test details (e.g., data splits, hyperparameters, how they were chosen, type of optimizer, etc.) necessary to understand the results?
    \item[] Answer: \answerYes{}  
    \item[] Justification: All training details have been included in the main paper and appendix.

\item {\bf Experiment statistical significance}
    \item[] Question: Does the paper report error bars suitably and correctly defined or other appropriate information about the statistical significance of the experiments?
    \item[] Answer:\answerNo{} 
    \item[] Justification: The paper reports performance metrics like MRR, Hit@1, and Hit@10 but does not include error bars or any discussion of statistical significance.

\item {\bf Experiments compute resources}
    \item[] Question: For each experiment, does the paper provide sufficient information on the computer resources (type of compute workers, memory, time of execution) needed to reproduce the experiments?
    \item[] Answer: \answerYes{} 
    \item[] Justification: Section~\ref{sec:Experiment details} states experiments used an Intel i9-14900H CPU without GPUs. Since MPRM has very small memory requirements, we did not report them.
    
\item {\bf Code of ethics}
    \item[] Question: Does the research conducted in the paper conform, in every respect, with the NeurIPS Code of Ethics \url{https://neurips.cc/public/EthicsGuidelines}?
    \item[] Answer:\answerYes{} 
    \item[] Justification: All datasets are translated versions of publicly available datasets. The paper does not appear to violate any ethical guidelines outlined in the NeurIPS Code of Ethics. It focuses on technical contributions without engaging in unethical practices such as data misuse or harmful applications.

\item {\bf Broader impacts}
    \item[] Question: Does the paper discuss both potential positive societal impacts and negative societal impacts of the work performed?
    \item[] Answer: \answerYes{} 
    \item[] Justification:  We discuss the impacts of MPRM in Section~\ref{sec:discuss}.
    
\item {\bf Safeguards}
    \item[] Question: Does the paper describe safeguards that have been put in place for responsible release of data or models that have a high risk for misuse (e.g., pretrained language models, image generators, or scraped datasets)?
    \item[] Answer: \answerNA{} 
    \item[] Justification: The paper poses no such risks.

\item {\bf Licenses for existing assets}
    \item[] Question: Are the creators or original owners of assets (e.g., code, data, models), used in the paper, properly credited and are the license and terms of use explicitly mentioned and properly respected?
    \item[] Answer: \answerYes{} 
    \item[] Justification: The paper cites the datasets and baseline methods used, which implies proper crediting.

\item {\bf New assets}
    \item[] Question: Are new assets introduced in the paper well documented and is the documentation provided alongside the assets?
    \item[] Answer: \answerYes{} 
    \item[] Justification: The novel algorithm is thoroughly described in Section 3.2 (Methodology) of the paper, including pseudocode (Algorithm 1), computational steps, and parameter design. 

\item {\bf Crowdsourcing and research with human subjects}
    \item[] Question: For crowdsourcing experiments and research with human subjects, does the paper include the full text of instructions given to participants and screenshots, if applicable, as well as details about compensation (if any)? 
    \item[] Answer: \answerNA{} 
    \item[] Justification: The paper involves no crowdsourcing or human subjects:

\item {\bf Institutional review board (IRB) approvals or equivalent for research with human subjects}
    \item[] Question: Does the paper describe potential risks incurred by study participants, whether such risks were disclosed to the subjects, and whether Institutional Review Board (IRB) approvals (or an equivalent approval/review based on the requirements of your country or institution) were obtained?
    \item[] Answer: \answerNA{} 
    \item[] Justification: No human subjects are involved.

\item {\bf Declaration of LLM usage}
    \item[] Question: Does the paper describe the usage of LLMs if it is an important, original, or non-standard component of the core methods in this research? Note that if the LLM is used only for writing, editing, or formatting purposes and does not impact the core methodology, scientific rigorousness, or originality of the research, declaration is not required.
    \item[] Answer: \answerNA{} 
    \item[] Justification: LLMs are only used for polishing text and understanding technical details.

\end{enumerate}

\end{document}